# SENTIMENT ANALYSIS FOR MODERN STANDARD ARABIC AND COLLOQUIAL

Hossam S. Ibrahim [1] , Sherif M. Abdou[2] and Mervat Gheith

[1] Computer Science Department, Institute of statistical studies and research (ISSR), Cairo University, EGYPT
[2] Information Technology Department, Faculty of Computers and information Cairo University, EGYPT

## ABSTRACT

*The rise of social media such as blogs and social networks has fueled interest in sentiment analysis. With the proliferation of reviews, ratings, recommendations and other forms of online expression, online opinion has turned into a kind of virtual currency for businesses looking to market their products, identify new opportunities and manage their reputations, therefore many are now looking to the field of sentiment analysis. In this paper, we present a feature-based sentence level approach for Arabic sentiment analysis. Our approach is using Arabic idioms/saying phrases lexicon as a key importance for improving the detection of the sentiment polarity in Arabic sentences as well as a number of novels and rich set of linguistically motivated features (contextual Intensifiers, contextual Shifter and negation handling), syntactic features for conflicting phrases which enhance the sentiment classification accuracy. Furthermore, we introduce an automatic expandable wide coverage polarity lexicon of Arabic sentiment words. The lexicon is built with gold-standard sentiment words as a seed which is manually collected and annotated and it expands and detects the sentiment orientation automatically of new sentiment words using synset aggregation technique and free online Arabic lexicons and thesauruses. Our data focus on modern standard Arabic (MSA) and Egyptian dialectal Arabic tweets and microblogs (hotel reservation, product reviews, etc.). The experimental results using our resources and techniques with SVM classifier indicate high performance levels, with accuracies of over 95%.*

## 1. INTRODUCTION

Much of the current research is now focusing on the area of sentiment analysis and opinion mining as the amount of internet documents increasing rapidly and where the Internet has become an area full of important information, views and meaningful discussions. These reasons make the statistical studies to evaluate the products, famous people and important topics are required and influential. Sentiment analysis refers to the use of natural language processing, text analysis and computational linguistics to identify and extract subjective information in the source materials. It aims to determine the attitude (judgment, evaluation or opinion) of a speaker or a writer with respect to some topic or the overall contextual polarity of a document (positive, negative, or neutral). In addition, Opinion mining can analyze opinions and extract other useful and complex information such as:  strength of the opinion polarity, opinion holder and opinion target.

There are mainly two main approaches for sentiment classification, namely: machine learning (ML) approach and semantic orientation (SO) approach. Machine learning is typically a supervised approach in which statistical machine learning algorithms is used for learning and testing data using machine learning classifiers such as Support Vector Machine (SVM), Naïve Bayesian Classifier, etc. [1-5]. Semantic orientation is an unsupervised approach in which linguistic rules derived from a language and sentiment lexicon are applied using rule-based classifiers for sentiment analysis[6]. Some challenges and problems exist during the study and research in this area, Pang and Lee [7] in their book, address all the difficulties and problems that face the opinion mining and sentiment classification. They discuss and show many features used in most researches such as: terms and their frequencies, Part-Of-Speech tag (POS), opinion words and phrases, syntactic dependency, negation, etc. to overcome most problems.

In this paper, we use a semi-supervised approach for sentiment analysis using a high coverage Arabic sentiment words lexicon which is automatically increased, and sentiment Arabic idioms/saying phrase lexicon to improve the classification process. We used support vector machine (SVM) classifier with linguistically and syntactically motivated features.

The remaining of the paper organized as follows: section2 related work of this area; section3 Arabic challenges; section4 describes the resources and lexicons built for this work; section5 explain the method and features used; section6 experimental setup and evaluation; and section7 the conclusion of our approach.

## 2. RELATED WORK

Sentiment words or phrases are the primarily used to express the writer's sentiment, emotion and opinion. Much of the work is focused on adjectives and adverbs, as they considered as the most obvious indication of the sentiments [5,8] and some others focus on verbs and nouns. Also, most of the work use (POS) tags to extract them from the text to build their sentiment lexicon [9].

Hatzivassiloglou and McKeown[6] used a method that automatically retrieves semantic orientation information using indirect information collected from a large corpus. Pang and Turney's approach [2,5] used supervised machine learning to classify movie reviews. Without classifying individual sentiment words or phrases. They achieved accuracies between 78.7% and 82.9%. Alexander and Patrick [10] use a corpus of 300000 text posts from a tweeter, collected and split automatically into three sets of text, they use emoticons to split the text into +ve / –ve emotions or objective (which contains no emoticons). Scheible and Schutze [11] present a novel graph-theoretic method for initial annotation of high-confidence training data for bootstrapping sentiment classifier. They introduce a new semi-supervised sentiment classifier that integrates lexicon induction with document classification. Their experiments are on customer reviews of multi-domains [12]. Davidiv and his group [3] use a supervised K-Nearest Neighbor (KNN) as a classifier. They work on different topics of English tweets. They use hash tags and smileys as features in the classification process. Also, Barbosa and Feng[4] applied their approach to tweets data and use an SVM classifier to classify the sentiment of tweets using abstract features. Abbasi[13]noted that one of the bottlenecks in applying supervised learning approach is the manual effort involved in annotating a large number of training data. Riloff and Wiebe[14,15] try to solve the problem by using a bootstrapping approach to label training data automatically. Their approach performed well, achieving 77% recall with 81% precision.

Most of the sentiment analysis work has focused on English language and European languages. Only a few works, try to solve the problems for morphologically rich languages such as Arabic language. Abdul-Mageed and Diab[16] present a newly labelled corpus of modern standard

Arabic (MSA) from the Newswire domain. The data are manually annotated for subjectivity and domain at the sentence level. Abdul-Mageed [17] presents a sentence-Level SSA system for Modern standard Arabic (MSA). They build a corpus of Newswire documents that manually annotated and an adjectives lexicon contains 3982 adjectives labelled as (positive, negative, neutral) for Newswire domain. They use language-independent features and MSA-Morphological features. They gained 71% of subjectivity and 95% of sentiment classification for news domain. Later on Abdul-Mageed[18] continue their work to Arabic social media content, including chat sessions, tweets, Wikipedia discussion pages, and online forums. They used author information (person vs. organization and gender) as features. In addition, they used Arabic specific features such as stemming, POS tagging, dialect and morphology features. They noted that POS tagging helps to improve classification process, they also noted that most dialectal Arabic tweets are subjective. Shoukry and Rafea[1] proposed an application for Arabic sentiment analysis by implementing a sentiment classification on 1000 Arabic tweets (500 positives and 500 negatives) using Machine Learning approach. They extracted all unigrams and bigrams terms from the corpus as a feature and calculate the frequency of each term, where the frequency of any term that exceeds 5 considered as candidate feature. The classification process performed using Weka Suite software. Their system got 72.8 % of success when using the combination. Mourad and Darwish[19] present in their work another way for (SSA) subjectivity and sentiment analysis on Arabic news articles and dialectal Arabic microblogs from twitter. They adopted a random graph walk approach to extend the existence Arabic SSA lexicon "ArabSenti" which performed by Abdul-Mageed [17] using Arabic-English phrase tables. They performed some experiments using Naïve Bayesian and SVM classifiers with some features (stem-level features, sentence-level features, and positive-negative emoticons). They got 80% accuracy for news domain and 72.5% accuracy for Arabic tweets as the best results.

In sentiment analysis, there are two main requirements are necessary to improve sentiment analysis effectively in any language and genres; high coverage sentiment adjectives lexicon and tagged corpora to train the sentiment classifier. So, Korayem [20] survey a different techniques and methods for building subjectivity and sentiment analysis systems for Arabic. They also review the most Arabic annotated corpuses built for Arabic sentiment analysis, such as: AWATIF which is a multi-genre, multi-dialect corpus for Arabic subjectivity and sentiment analysis [21]. It is collected from three different resources: Penn Arabic Treebank (PATB) which is a collection of news wire topics of different domains, Wikipedia user talk pages and a user conversation on web forum sites. Opinion corpus for Arabic (OCA) is built by Rushdi-Saleh[22]. The data is collected from several blogs of movie reviews, obtaining a total of 500 comments (250 positive and 250 negative). They used two machine learning classifiers (SVM, NB) trained in determining the polarity of a review and to compare the corpus performance. Elarnaoty [23] presented MPQA subjective lexicon and Arabic opinion holder corpus. They collected 150MB news articles from news websites, only 1MB only of their corpus is pre-processed, including sentence segmentation, Morphological analysis, part of speech tagging (POST), semantic analysis, Named Entities Recognition (NER), subjective analysis, and manual annotation for opinion holders.

Abdul-Mageed and Diab [24] present SANA which is a large scale multi-Genre multi-Dialect lexicon for Arabic SSA. In addition to Modern standard Arabic, SANA covers also, Egyptian Dialect and Levantine dialect Arabic. It developed using both manual and semi-automatic techniques. SANA is a huge Arabic lexicon, but it's still under construction and it is not yet fully applied to SSA tasks. Eshrag and Verena [25] presented a newly collected data set of 8868 gold-standard Arabic tweets. The corpus is collected from twitter API and annotated manually for subjectivity and sentiment analysis (SSA). For the classification process, they used a rich set of linguistically motivated features (morphological, syntactic and semantic features and ArabiSenti[17] lexicon. In addition, they present a lexicon of 4,422 annotated Arabic sentiment words and phrases. They noted that the Arabic twitter train and test sets will be released via the ELRA data repository.

Most of the recent work in Arabic languages is not yet releasing their resources and some of them sharing the same weak points such as; using a few features for sentiment classification process. Most systems can't handle negation in the statement which is very important as in [1,19]. Also, the redundancy in the training data causes an ambiguity in sentiment, and the lack or nonexistence of Egyptian dialect list affects the accuracy of the classification results as in [1]. Some of them work as domain-dependent for Newswire or Arabic MSA only and exclude dialects and its complexity as in[17].

Our work shows that the using of tailored wide coverage Arabic sentiment polarity lexicon and tailored Arabic sentiment idioms-saying lexicon and employing rich set of linguistically motivated features (linguistic, Sentence-level features and Syntactic features) improve the classification process which affect the system performance. The sentiment polarity lexicon is automatically expanded and automatically detects the sentiment orientation of new sentiment words using POS tags, synset aggregation technique and free online Arabic dictionaries, synonyms-Antonyms lexicon. We also measure the impact of using an expansion technique on classification results and show that using tailored idioms-saying Arabic lexicon improves the classification performance.

## 3. ARABIC LANGUAGE AND CHALLENGES

Arabic is one of the six official languages of the United Nations. According to Egyptian Demographic Center[1], it is the mother tongue of about 300 million people (22 countries). There are about 135 million Arabic internet users until 2013. The orientation of writing is from right to left and the Arabic alphabet consists of 28 letters. The Arabic alphabet can be extended to ninety elements by writing additional shapes, marks, and vowels.

Most Arabic words are morphologically derived from a list of roots that are tri, quad, or pent-literal. Most of these roots are tri-literal. Arabic words are classified into three main parts of speech, namely nouns, including adjectives and adverbs, verbs, and particles. In formal writing, Arabic sentences are often delimited by commas and periods. Arabic language has two main forms: Standard Arabic and Dialectal Arabic. Standard Arabic includes Classical Arabic (CA) and Modern Standard Arabic (MSA) while Dialectal Arabic includes all forms of currently spoken Arabic in day life and it vary among countries and deviate from the Standard Arabic to some extent [26]. Modern standard Arabic (MSA) considered as the standard that commonly used in books, newspapers, news broadcast, formal speeches, movies subtitles, etc. of the Arabic countries.

This research is concerned about modern standard Arabic (MSA) and the Egyptian dialect. The complexity here is that all the natural language processing approaches that has been applied to most languages, is not valid for applying on Arabic language directly. The text needs much manipulation and pre-processing before applying these methods. The main challenging aspects of sentiment analysis and opinion mining exist with use of other types of words, sentiment lexicon construction, dealing with negation, degrees of sentiment, complexity of the sentence / document, words in different contexts, etc.

---

[1] Internet world stats "usage and population statistics", http://www.internetworldstats.com/
[1] Twitter search API:
http://search.twitter.com/search.atom?lang=ar&rpp=100&page={0}&q={1}

# 4. DATA SET AND ANNOTATION

## 4.1. Corpus

We built our own corpus contains 2000 Arabic sentiment statements includes 1000 MSA tweets, Arabic dialect tweets and 1000 microblogs (hotel reservation comments, product reviews, TV program& movie comments). The corpus includes data in both Modern Standard Arabic and Egyptian dialectal Arabic. We started collecting data from June 2012 using Twitter API[2] and different microblogs and forum websites such as: http://www.booking.com/, http://forums.fatakat.com , http://ejabat.google.com , etc.

We collect about 10 thousand Arabic tweets and 10 thousand Arabic comments and reviews. The selected data is performed according to specific conditions; Sentiment, hold one opinion, MSA, Egyptian dialect, not sarcastic, no insult, subjective. The data is selected and annotated manually as; positive and negative sentiment using three raters (native speakers) of age between 35 and 45. The inter-annotator agreement in terms of Kappa (K) is measured, it's about 96%. Kappa determines the quality of a set of annotations by evaluating the agreement between annotators. The current standard metric used for measuring inter-annotator agreement on classification tasks is the Cohen Kappa statistic [27].

## 4.2. Sentiment Lexicon

Our approach is basically lexicon-based in that it determines the polarity value of terms by matching with lexical known polarity terms that in the polarity lexicon. In this work, we built two lexicons; Arabic sentiment words lexicon and Arabic sentiment idioms/saying phrase lexicon. Most of the work use adjectives only for sentiment analysis, and some of them use nouns, verbs, adverbs or a combination of them. We decided to use sentiment adjectives and nouns of adjectives because we noticed that some opinion statements do not include any adjectives, but express negative or positive sentiment such as; "ما حدث في بلادنا يعطي شعور بالكراهية و الحقد" means "What happened in our country gives a feeling of hatred and malice". There aren't any adjectives in this example, although it expresses the negative sentiment, the words "الكراهية"-"hatred", "الحقد"-"malice" is the noun of adjectives "مكروه"-"Hateful", "حقود"-"malicious" which express the negative sentiment in the example.

We introduce a (5244) sentiment adjectives lexicon "ArSeLEX" which is manually created and automatically expandable. Starting with the gold-standard 400 adjectives collected manually as a seed from different website[3] specialist in Arabic language and grammar. The lexicon is expanded manually for the first time by collecting synonyms and antonyms of each word using different Arabic dictionaries and label each word with one of the following tags [negative (NG), positive (PO), neutral (NU)].  We applied an algorithm to expand and detect the sentiment orientation automatically of new sentiment words using synset aggregation technique and free free online Arabic dictionaries and lexicons such as Google translation API and free online Arabic synonyms and antonyms thesaurus[4] and calculate the term frequency (TF) of each sentiment word from our corpus. The expansion algorithm will be described in details in section (5.1).

---

[3]  http://www.languageguide.org/arabic/grammar/adjectives.jsp  , http://en.wiktionary.org/wiki/Category:Arabic_adjectives  , http://arabic.speak7.com/arabic_vocabulary_adjectives.htm .
[4] http://translation.google.com,
 http://www.almaany.com/thesaurus.php?language=arabic.

**4.3. Idioms phrase Lexicon**

Since we are working on the Arabic language and Egyptian colloquial and most internet users who write in these languages are expressing their opinions and feelings with sarcastic way, old wisdom, old saying and idioms, which exceeded more than 20% of the total opinions. The sarcastic sentences are excluded because it is hard to deal with, e.g., "What a great car! It stopped working in two days." Sarcasms are not so common in consumer reviews about products and services, but are very common in political discussions, which make political opinions hard to deal with[28]. But, there are a lot of phrases which represent old wisdoms and popular idioms that people used to use it in their comments to represent their opinion directly as shown in Table 1.

Table1. Examples of old saying or idioms

| Example1 | "تسليم السلطة للبرلمان تعني تسليم القط مفتاح الكرار" |
| | tslym AlslTp llbrlmAn tEny tslym AlqT mftAH AlkrAr |
| | The handover of power to the parliament means delivering the key of Karar to the cat |
| Example2 | "بعد 30-عاما في الحكم أدركنا إن إلي افتكرناه موسي طلع فرعون" |
| | bEd 30 EAmA fy AlHkm >drknA <n <ly AftkrnAh mwsy TlE frEwn |
| | "After 30 years in power, we realized that what we thought he was Moussa became Pharaoh" |

These phrases cannot be neglected, it gives different meanings when it segmented or translated by dictionaries, therefore, there must be a glossary or lexicon covering these terms and phrases. The absence of these lexicons pushes us to build a phrase lexicon contains sentiment wisdom, idioms and aphorisms for Egyptian dialect. A (12785) wisdoms and idioms are collected from websites and books that specialized in this task, such as [29-33]. We select (3296) phrases which are sentiment and commonly used, and annotate it manually as negative or positive. We are sure that this lexicon will improve the sentiment classification process, and this is what already been achieved.

In the above examples, there are not any adjectives, and the classifiers will labelled them as neutral sentiment although they express a negative sentiment. It is clear in the phrase " تسليم القط مفتاح الكرار" of the first example which means "deliver the key of Karar to the cat" when translated by dictionaries in spite of it means "Give the thief the key of the safe" which expresses negative sentiment, also the phrase "افتكرناه موسي طلع فرعون" of the second example, which means "we thought he was Moussa became Pharaoh" when translated by dictionaries, although it means, "what we thought that he was a good man, he becomes a very bad man" which expresses negative sentiment. we applied a heuristic rule to the topic that contains idioms to prevent redundancy in classification process of sentiment analysis systems. Specifically, we replace known idioms and proverbs with text masks. For example, the idiom "crocodile tears", known to have negative (NG) sentiment polarity, should be replaced by (NG_Phrase), similarly replaced (PO_Phrase) by the idiom or proverbs that have positive sentiment. This step is very helpful in sentiment polarity detection process of the sentiment analysis and opinion mining systems. For example in un-supervised approaches a range of sentiment values from -3 to +3 assigned for idioms and a range of sentiment values from -1 to +1 assigned for sentiment words etc.. the net polarity of the topic can be obtained from the sign of the net score produced from all assigned values, if the net score is negative then the sentiment polarity is (NG) and vice versa. For supervised approaches it can be used in the training process, to learn the classifier that the existence of these masks increases the negativity or positivity of the topic.

# 5. SENTIMENT ANALYSIS

The system consists of two parts; part1 includes the pre-processing and lexicon expansion, part2 includes the features extraction and classification. The pre-processing includes data cleaning (remove foreign characters, symbols, numbers, etc.), remove stopwords and normalization (unified Arabic characters and removes all diacritics such as ['إ', 'أ', 'آ' => 'ا'], ['ة', 'ه' => 'ه'], ['ي', 'ى' => 'ي']).

## 5.1. Automatic expansion and orientation detection

We considered our polarity lexicon which is manually collected and annotated as the basic lexicon. It consists of 5244 annotated sentiment words includes (2003 Positive (PO), 2829 negative (NG), 412 neutral (NU)). In this research, we propose a simple and yet effective method of utilizing the synonym set and antonym set extracted from free online Arabic dictionaries and lexicons to predict the sentiment orientations of sentiment words.

In general, adjectives share the same orientation as their synonyms and opposite orientations as their antonyms [8]. We followed Hu and Liu by using this idea to predict the orientation of the sentiment word. So, we built a method consists of four steps as shown in Figure 1. First the method use AMIRA part of speech (POS) tag [34] to extract the expected words that can be a sentiment words such as Adjectives (JJ), Nouns (NN) and verbs (VB). Second, remove redundant sentiment words by deleting all repeated words that exist in the basic lexicon. Third, send each word of the remaining words to online Arabic dictionary and lexicons to get translation and all its synonyms.

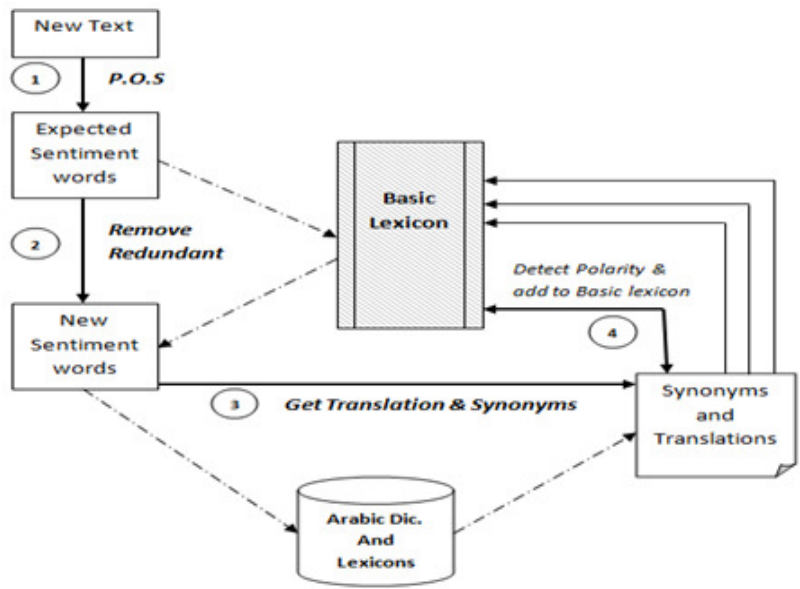

Figure 1. Expansion Flow Diagram

Finally, the synset of the given word and the antonym set are searched. If a synonym/antonym has known orientation, then the orientation of the given word could be set correspondingly. So, we have three cases:

1. If we received a translation of the word and its synonyms, and the synonyms found in the basic lexicon which labelled as positive, then the polarity of the word is (PO), or (NG) polarity if the word related to negative class. The word adds to the basic lexicon automatically with its translation and polarity. Example as shown in Table2:

Table 1. Example of 1st case in expansion method

| Arabic sentiment word | BuckWalter Transliteration | English Translation | Arabic Synonyms (syn.) | Synonyms English Translation | (syn.) Polarity | Expected Polarity |
|---|---|---|---|---|---|---|
| مسرور | Msrwr | Delighted | فرحان سعيد مبتهج | Pleased Happy Glad | PO PO PO | PO |

2. If we received a translation and its synonyms, but the synonyms have different polarities. The word considered as a conflict of synonyms (COS), labelled as neutral (NU) and didn't add to the basic lexicon. Example as shown in Table3:

Table 2. Example of 2ed case in expansion method

| Arabic sentiment word | BuckWalter Transliteration | English Translation | Arabic Synonyms (syn.) | Synonyms English Translation | (syn.) Polarity | Expected Polarity |
|---|---|---|---|---|---|---|
| شديد | $dyd | Intense | قوي عنيف حاد | strong violent keen | PO NG PO | NU |

3. If there is no translation and no synonyms of the word. The word considered as out of vocabulary (OOV), in most cases it is a dialect word. So, it adds automatically to the lexicon if the user agreed that is a sentiment dialect word and enter its polarity or it adds to prevent list if the word is not sentiment word. Example as shown in Table4:

Table 3. Example of 3ed case in expansion method

| Arabic sentiment word | BuckWalter Transliteration | English Translation | Arabic nearest meanings | English Translation | (syn.) Polarity | Expected Polarity |
|---|---|---|---|---|---|---|
| هايف | hAyf | No Translation (dialect) | تافه ابله مستهتر | Fiddling Idiot Playboy | NG NG NG | NG |

We experiment the impact of this lexicon expansion and it shows that it have a measurable impact on the system performance and accuracy which will be clarified in the results of our experiments in the section6. This lexicon is built for MSA and Egyptian dialect as we focus in this work and we intend to increase this lexicon to be able for the all the six Arabic dialects [35], and introduce it as a resource for multi-genre, multi-dialects sentiment lexicon. The lexicon consists of five columns, namely: Arabic sentiment word, English gloss, Buckwalter transliteration, sentiment polarity and term frequency (TF), for more information about our lexicon "ArSeLEX" see[36].

## 5.1. Features Selection and Extraction

Selection of optimal features set is considered as the most challenging aspects for any machine learning approach in NLP area. We employed different features to our sentiment analysis system such as standard features, sentence-level features, linguistic features, etc. We built an algorithm for selection and extraction features from text and construct a one-line statement (SVM format) for each topic. The topic here can be one tweet, review or comment which consists of one or more statements.

### 5.1.1. Standard Features

For sentiment classification, we apply two main binary features for sentiment words, has_PO_senti and has_NG_senti which determine if the word has positive or negative sentiment polarity using our basic lexicon. And two main binary features for idiom phrases, has_PO_ph and has_NG_ph which define if the topic contains any idiom/saying phrases and if it's positive or negative using our idiom/saying lexicon. Other standard features which are simply counting the number of positive (W_PO), negative (W_NG), neutral (W_NU) sentiment words in the topic.

### 5.1.2. Sentence-Level Features

Long sentence can heavily affect the classification accuracy. Pang and Lee[2] noted this problem in their work in products review classification is when the reviewer starts by many negative sentences, but at the end negates the whole review with a short sentence of "but I liked it!" [37], to overcome this problem, we use two features; (A) term frequency (TF). The term frequency can solve the part of the problem by increasing the polarity strength of the polar word. (B) Polar word position. In fact, the position of a token within a textual unit (e.g., in the middle vs. near the end of a document) can potentially have important effects on how much that token affects the overall sentiment. The adjective term that appears in the beginning of the sentence increase the polarity strength than the one that in late position [2]. For example " هذا المسلسل رائع لكن يوجد ملل في بعض حلقاته", "This series is wonderful, but there is boredom in some episodes". The sentiment classification of this example when using the polarity lexicon only will be neutral due to the appearance of two (positive and negative) adjectives which in fact it is a positive sentiment because the first impression of the writer is "This series is wonderful…". This result will be achieved when using the word position feature. So, we apply two features PO_W_Position and NG_W_Position. Each feature is increased by the value of position of the polar word in the sentence which calculated by (number of sentence words / word position). In addition, we apply a number of words in the sentence No_of_words as a feature to improve the classification accuracy against long statements.

### 5.1.3. Linguistic Features

We explored a wide variety of linguistic and lexical features that can accurately handle complex language structures in the Arabic language such as negation, intensifiers, etc. In this work we used the following set of features:

- **Contextual Shifter (Negation)**

    Handling negation can be an important concern in opinion and sentiment analysis. For example "أنا احب هذا الكتاب" "I like this book" and "انا لا احب هذا الكتاب" "I don't like this book", the two sentences considered to be very similar by most commonly-used similarity measures, the only different is the negation term, forces the second sentences into opposite polarity. So,

there is a list of negation terms in Arabic language (MAS) and Egyptian dialect such as "لا, مش, ليس لأن,..." which can change the sentiment polarity of terms from positive to negative and vice versa when any of them occurred before the sentiment word. Our classifier can handle this negation when occur and opposite the sentiment word polarity. Also, we employed two features, namely: Is_Negation a binary feature indicates that there is a negation in the sentence and N_O_Negation feature to express how many negation terms appear in the sentences.

– **Contextual Intensifiers**

Intensifier terms such as 'too' in 'too difficult' and 'very' in 'be very pleased' act to strengthen the polarity value of the sentiment term. There is a list of intensifier terms in Arabic language (MAS) such as "جدا, اوي, بشدة,...." which written after the sentiment word to strength its sentiment polarity. In our approach we calculate the effect of intensifier terms by doubling the base value of a sentiment term. For example "هذه المرأة جميلة اوي" "This woman is very beautiful" and "هذا المكان قذر جدا" "This place is very dirty". In example1 the adjective 'جميلة' 'beautiful' is judged as positive by matching with the polarity lexicon, so it receives the positive value +1 and the contextual intensifier 'اوي' 'very' increases the base value of the adjective to +2 in order to emphasize positive polarity in this sentence. Where, in example2 the contextual intensifier 'جدا' 'very' increases the base value of the adjective to -2 in order to emphasize negative polarity of the sentiment word.

– **Questions**

The questions are generally meant to search for something you need or do not exist. In sentiment dialogues and opinion rating, questions have the same meaning, but express negative sentiment because the writer feels sad or angry or wonder because of the lack of what it needs from his point of view. This is what people consider an effective way or method for Denunciation, rejection and condemnation. Therefore, we consider the existence of the question in a sentence gives negative sentiment polarity in most of the sentences that we observed. For example: "أين الحرية و العدالة التي كنا نحلم و نطالب بها؟" "Where are the freedom and justice that we dreamed and asked for?", "هل يوجد راجل محترم في الحكومة الجديدة؟ انا شخصيا مش شايف حد" "Is there a respectable person in the new government? Personally, I do not see anyone". So, we employed two features, namely: Is_Question a binary feature indicates that there is a questionable term in the sentence and N_O_Question feature to express how many question terms appear in the sentences. The classifier detects it using a list of question terms exist in Arabic language (MAS).

– **Supplication and Wishful**

In fact, supplication is a matter of asking GOD for help. People always asking GOD for blessing someone, revenge from others, helping themselves, etc. They want something desperately and pray for it or wish it. So, it expresses negative sentiment. In our countries, people always use this style of talk. For example: "اتمني رجوع مصر الي حكم الملكية بلا جمهوريات بلا جوع بلا بلطجة و تعب" "I hope you return to the rule of Egypt's property without republics without hunger and fatigue without bullying", "يارب يحكمنا رئيس يحكم بالعدل و المساواة" "Lord we need a president govern with justice and equality". Therefore, we employed two features, namely: Is_wishful a binary feature indicates that there is a supplication or wishful term in the sentence and N_O_wishful feature to express how many supplication or wishful terms appear in the sentences. Our classifier detects it using a list of supplication or wishful terms exist in Arabic language (MAS).

**5.1.4. Syntactic features for conflicting phrases**

Syntactic features include phrase patterns, which make use of POS tag n-gram patterns to detect the inflection phrases[13]. Following the authors [38,39] they noted that phrase patterns such as "n+aj" (noun followed by positive adjective) typically represent positive sentiment orientation, while "n+dj" (noun followed by negative adjective) often express negative sentiment. In this work we used POS tag and bi-gram to extract conflicting phrases patterns consist of two sentiment words that opposite in polarity; noun followed by adjective or adjective followed by noun, one of them is positive polarity and the other is negative polarity. For example the pattern "خدمة سيئة" "Bad duty/favor" consists of two sentiment words, the noun 'خدمة' 'duty or favor' which express positive polarity and the adjective 'سيئة' 'Bad' which express negative polarity. The net score of their polarity is 0 when using polarity lexicon only, but our classifier can detect these phrases and labeled it as negative polarity. As we noticed in most cases, if there exists a pattern of two opposite sentiment words, it yields a negative sentiment polarity such as "خيانة الأمانة" "Dishonesty or (Betrayal of Municipality)", "فساد أخلاقي" "moral corruption", "قلة أدب" "incivility or (little propriety)" and "سعادة وهمية" "fake happiness" etc.

## 5. EXPERIMENTAL RESULTS AND EVALUATION

In this study, we use a Support Vector Machine (SVM) light [40] for classification, as SVMs are known for being able to handle large feature spaces. We experiment with various kernels and parameters settings and find that linear kernels yield the best performance for our work.

We run two sets of experiments. In the first set we divide the data into 80% for training and 10% for developing and 10% for testing. The classifier and features are optimized during the developing set and all the results that we report are in the test set. Also, we report accuracy as well as precision, recall and F-measure for each experiment. F-measure is calculated by the following equation.

$$F - Measure = 2 \cdot \frac{precision \cdot recall}{precision + recall}$$

Our corpus (2000 topics) consists of four types of data as follows (997 tweets 49%, 300 hotel reservation comments 16%, 346 product reviews 17%, 357 TV-Programs comments 18%). The corpus includes (971 negative topics (NG), 1029 positive topics (PO)). In test set experiments we use 10% of data (200 topics) which consists of 99 tweets, 30 hotel reservation comments, 35 product reviews and 36 TV program comments respectively to the same proportions of the four types of data in the corpus. We experiment on each type of data separately to measure the effect of complexity of each type on classification accuracy and one experiment to overall combined data as shown in Table 5.

In the same set, we performed the same experiments again after applying the expansion algorithm of our lexicon on the same data to measure the impact of using an expansion technique on classification results as shown in Table 6. The expansion technique extracts and adds a new sentiment words to the polarity lexicon automatically.

In the second set, we considered the results of the first set experiments as Baseline results and we collect four new collection of data (400 topics) consists of 100 tweets, 100 hotel reservation comments, 100 product reviews and 100 TV program comments. We perform the two groups of experiments (before expansion and after expansion) on each type to re-test the system and ensure the efficiency of the system and measure the expected increase in the classification accuracy results. We report the experiments of the best results as shown in Table 7 and Table 8.

As shown in the tables 5,6,7 and 8 the final results show that our sentiment analysis system yields high performance and efficiency in sentiment classification of the types of data that applied to the system. Also, tweets are considered as the most complex type of data that can be analyzed sentimentally due to the sarcastic way which always used to express the writer's opinions. We noticed that more than 10% of tweets are written in sarcastic ways which are very complex to handle with regular statistical methods. On the other hand the other types of data give good results in sentiment analysis because of the direct way that used to express opinions in product evaluation.

Table 5. Baseline results for Arabic sentiment classification before lexicon expansion

| Data | SVM classifier | | | |
|---|---|---|---|---|
| | Accuracy | Precision | Recall | F-Measure |
| Tweets | 80.95238% | 84.94624% | 84.94624% | 84.9462% |
| Hotel res. | 94.63087% | 98.26087% | 94.95798% | 96.5812% |
| Product rev. | 93.84615% | 96.22642% | 96.22642% | 96.2264% |
| TV prog. Comm. | 88.40580% | 92.10526% | 87.50000% | 89.7436% |
| Total data | 89.06977% | 93.31104% | 91.17647% | 92.2314% |

Table 6. Baseline results for Arabic sentiment classification after lexicon expansion

| Data | SVM classifier | | | |
|---|---|---|---|---|
| | Accuracy | Precision | Recall | F-Measure |
| Tweets | 83.67347% | 87.09677% | 87.09677% | 87.0968 |
| Hotel res. | 95.30201% | 98.27586% | 95.79832% | 97.0213 |
| Product rev. | 95.38462% | 98.11321% | 96.29630% | 97.1963 |
| TV prog. Comm. | 94.20290% | 97.43590% | 92.68293% | 95.0000 |
| Total data | 90.46512% | 94.31438% | 92.15686% | 93.2231 |

Table 7. Results for Arabic sentiment classification before lexicon expansion

| Data | SVM classifier | | | |
|---|---|---|---|---|
| | Accuracy | Precision | Recall | F-Measure |
| Tweets | 87.50000% | 84.1804% | 100.00000% | 91.41085 |
| Hotel res. | 93.54839% | 95.83333% | 95.83333% | 95.83333 |
| Product rev. | 96.66667% | 95.65217% | 100.00000% | 97.77777 |
| TV prog. Comm. | 93.33333% | 93.33333% | 93.33333% | 93.33333 |
| Total data | 94.30894% | 91.78082% | 98.52941% | 95.03545 |

Table 8. Results for Arabic sentiment classification after lexicon expansion

| Data | SVM classifier | | | |
|---|---|---|---|---|
| | Accuracy | Precision | Recall | F-Measure |
| Tweets | 90.62500% | 88.00000% | 100.00000% | 93.61702 |
| Hotel res. | 96.77419% | 96.00000% | 100.00000% | 97.95918 |
| Product rev. | 100.00000% | 100.00000% | 100.00000% | 100 |
| TV prog. Comm. | 96.66667% | 100.00000% | 93.33333% | 96.55172 |
| Total data | 95.12195% | 93.15068% | 98.55072% | 95.77464 |

Results in Tables 6 and 8 shows that the lexicon expansion had a large effect on sentiment classification with improved accuracy, precision, and recall with improvements ranging between 1-4 % (absolute) for all types of data. Also, tables 5 and 7 reported that the coverage of the polarity lexicon increases gradually.

In this work we reported some difficulties that we intend to handle in the future, such as negation can often be expressed in rather subtle ways, Sarcasm and irony can be quite difficult to detect, some tweets or comments may contain dual or comparing opinions, and English transliteration text which called (Franko-Arab text or Roman Text), some users write English statements transliterated with Arabic letters such as "I love him" to "اي لوف هيم" instead of "انا احبه" in Arabic language or "Valentine's day" to "فالنتاين" instead of "عيد الحب"

# 7. CONCLUSIONS

In this paper, we presented a sentiment analysis system for MSA and Egyptian dialect using a corpus of different types of data (tweets, product reviews, TV program comments and Hotel reservation). We employed a number of novels and rich feature sets to handle the valence shifters (negation, intensifiers), question and supplication terms and to improve the classification performance. We show that exploiting idioms and saying lexicon with a high coverage polarity lexicon has the largest impact on classification accuracy. Also, the automatic expansion of the polarity lexicon yields a great effect on sentiment classification. The results obtained show that our SA system is very promising. For the future we plan to increase our Egyptian dialect polarity lexicon and the Egyptian idioms/saying lexicon, since to the best of our knowledge, there is no resource of any of them are available and we will make that lexicons available to the community at large.